\pdfoutput=1

\documentclass[11pt]{article}

\usepackage[final]{acl}
\usepackage{diagbox}
\usepackage{times}
\usepackage{latexsym}

\usepackage[T1]{fontenc}

\usepackage[utf8]{inputenc}

\usepackage{microtype}

\usepackage{inconsolata}

\usepackage{graphicx}

\usepackage[most]{tcolorbox}
\usepackage{microtype}
\usepackage{hyperref}
\usepackage{url}
\usepackage{booktabs}
\usepackage{amsmath} 
\usepackage{lineno}
\tcbset{
  promptbox/.style={
    colback=gray!20,
    colframe=black!60,
    boxrule=0.5pt,
    arc=2mm,
    outer arc=2mm,
    boxsep=4pt,
    left=6pt,
    right=6pt,
    top=4pt,
    bottom=4pt,
    fonttitle=\bfseries,
    title=Prompt
  }
}
\usepackage{enumitem}
\usepackage{booktabs}
\usepackage{graphicx}
\usepackage{multirow, makecell}
\usepackage{soul}        
\usepackage{xcolor}      
\usepackage{amssymb}

\usepackage{times}
\usepackage{latexsym}
\usepackage{booktabs}
\usepackage{color, soul}
\usepackage[T1]{fontenc}

\usepackage[utf8]{inputenc}
\usepackage[inkscapelatex=false]{svg}
\usepackage{microtype}

\usepackage{inconsolata}
\usepackage{multirow, makecell}

\usepackage{graphicx}
\usepackage{amsmath}

\usepackage{graphicx}
\usepackage{subcaption}

\newcommand{\hlpink}[1]{%
  {\sethlcolor{red!20}\hl{#1}}%
}
\newcommand{\hlgreen}[1]{%
  {\sethlcolor{green!20}\hl{#1}}%
}

\definecolor{darkblue}{rgb}{0, 0, 0.5}
\hypersetup{colorlinks=true, citecolor=darkblue, linkcolor=darkblue, urlcolor=darkblue}

\newcommand{\ours}{\texttt{ACPO}}
\newcommand{\acpo}{\texttt{ACPO}}
\newcommand{\acpofive}{\texttt{ACPO(5,5)}}
\newcommand{\factalign}{\texttt{FactAlign}}
\newcommand{\llamat}{\texttt{Llama-3-8B-Instruct}}
\newcommand{\phit}{\texttt{Phi-3-mini-4k-instruct}}
\newcommand{\rawmodel}{\texttt{RawModel}}

\title{Atomic Consistency Preference Optimization for Long-Form Question Answering}

\author{
  Jingfeng Chen\thanks{\ \ Equal contribution.} \\
  \texttt{jingfenc@andrew.cmu.edu}
  \And
  Raghuveer Thirukovalluru\footnotemark[1] \\
  \texttt{raghuveer.thirukovalluru@duke.edu}
  \AND
  Junlin Wang \\
  \texttt{junlin.wang2@duke.edu}
  \And
  Luo Kaiwei\\
  \texttt{luokw1@chinatelecom.cn}
  \And
  Bhuwan Dhingra\\
  \texttt{bhuwan.dhingra@duke.edu}
}

\begin{document}\maketitle
\begin{abstract}
Large Language Models (LLMs) often produce factoid hallucinations - plausible yet incorrect answers. A common mitigation strategy is model alignment, which improves factual accuracy by training on curated (factual, non-factual) pairs. However, this approach often relies on a stronger model (e.g., GPT-4) or an external knowledge base to assess factual correctness that may not always be accessible. Addressing this, we propose Atomic Consistency Preference Optimization (\acpo), a self-supervised preference-tuning method that enhances factual accuracy without external supervision. \acpo\ leverages atomic consistency signals (i.e., the agreement of individual facts across multiple stochastic responses) to identify high- and low-quality data pairs for model alignment. Despite being fully self-supervised, \acpo\ outperforms the strong supervised alignment baseline by $1.95$ points averaged across Phi-3 and Llama3 on the LongFact and BioGen datasets, demonstrating its effectiveness in improving factual reliability without relying on external models or knowledge bases.
\end{abstract}




\section{Introduction}



Large Language Models (LLMs) have emerged as powerful tools for accessing information through natural language generation. Long-form factoid question-answering (QA), in particular, plays a crucial role in human interactions with LLMs for information retrieval \citep{alkhamissi2022review}. However, a significant concern with LLMs is their tendency to produce content that appears plausible but is factually incorrect, a phenomenon commonly referred to as hallucination \citep{rawte2023troubling,xu2024wizardlm,huang2025survey}. This issue is especially critical in the use of LLMs in domains like medical diagnosing, news reporting, and educational tutoring. To mitigate this issue, numerous strategies have been proposed.

One of the common ways to mitigate hallucinations and improve factuality is model alignment. Model alignment for factuality leverages curated (factual, non-factual) data pairs to align the model toward generating more factual content \citep{zhang2024self, tian2023fine}. Typically, these data pairs are identified using a retriever paired with a knowledge base or a more advanced language model (like GPT-4) \citep{huang2024factalign, zhang2024self}. However, the applicability of these techniques is often limited by two key factors.

First, the unavailability of structured knowledge bases in many scenarios, particularly in low-resource domains such as IT technical support \citep{yang-etal-2023-empower}, medicine, and law \citep{sengupta-etal-2025-exploring} restricts the effectiveness of these methods. Second, relying on advanced proprietary APIs (e.g., GPT-4 or Gemini 2.5) to score alignment data is very expensive. It also introduces serious privacy risks, especially when scored data includes sensitive information. These challenges underscore the need for self-supervised approaches that can enhance factual accuracy and reduce hallucinations without relying on knowledge bases or external models.


Conversely, several inference-time techniques, such as ASC \citep{thirukovalluru2024atomicASC}, CoVe \citep{dhuliawala2024chain}, and USC \citep{chen2024universal}, operate in a fully self-supervised manner (not relying on any external models/knowledge bases) by evaluating an LLM's output using \textit{self-consistency} or \textit{self-evaluation} based mechanisms. These methods are designed to identify and eliminate non-factual components, delivering a more reliable final response. However, these methods are computationally expensive at inference time, often necessitating multiple LLM calls (e.g., verifying each individual fact) to improve performance.

Inspired by the success of these inference-time hallucination reduction techniques in long-form question answering, we propose a self-supervised training approach that extends these principles to the alignment phase, enabling models to learn inherent factuality without reliance on external supervision. 

Although versions of self-supervised preference tuning has been explored in prior works, such as FactTune and SKT \citep{zhang2024self, tian2023fine}, these methods remain computationally expensive due to their heavy reliance on GPT-3.5 for extracting individual atomic facts and verification questions. Further, for factual calibration, self-consistency-based approaches have been shown to outperform confidence estimation methods such as self-evaluation in SKT and atomic question confidence in FactTune \citep{huang2024calibrating}. Recent work on reasoning tasks further highlights that self-consistency-based alignment tuning outperforms alternative methods \citep{prasad2024self}.

We propose Atomic Consistency Preference Optimization (\acpo), a scalable self-supervised framework for enhancing factuality in long-form generation. Unlike prior methods, \acpo\ does not depend on external knowledge bases or stronger models. Instead, it relies solely on a base large language model and a lightweight BERT-based embedder. Specifically, \acpo\ applies atomic self-consistency—the factual agreement across multiple stochastically sampled responses \citep{thirukovalluru2024atomicASC}—to efficiently construct preference-alignment pairs without supervision. Our contributions are:
\begin{itemize}[noitemsep, topsep=0pt, leftmargin=*]
\item \acpo\, a novel, privacy-guaranteed, cost-efficient self-supervised preference tuning method to improve long-form factoid QA abilities without reliance on any stronger LLMs or knowledge bases.

\item \acpo\ effectively reduces hallucinations and outperforms \factalign, a strong supervised baseline, by improving factual precision by an average of $+1.95$ points on Phi-3 and Llama3 over fact-checking benchmarks: LongFact \citep{wei2024long} and BioGen \citep{min2023factscore}.

\item Through systematic ablations, we show that atomic self-consistency provides a strong and effective signal for the reinforcement learning step of large language models, outperforming its direct application at inference time. 
\end{itemize}

\section{Related Work}
This section provides a comprehensive overview of inference‑time methods and preference‑tuning approaches aimed at reducing hallucinations and improving long‑form question answering.

\subsection{Inference Time Methods}
\subsubsection{Using Retrievers, Self-Evaluation}

FactScore \citep{min2023factscore} uses an external retriever to evaluate and improve response factuality. LongFact \citep{wei2024long} extends the original FactScore metric by incorporating an F1-based evaluation for recall level factual assessment. Chain of Verification (CoVe) \citep{dhuliawala2024chain} introduces a method that generates multiple verification questions for a given response, retaining only the segments that can be independently verified. Similarly, \citet{agrawal2024language} filters non-factual content from list-style answers using indirect self-evaluation questions.


\subsubsection{Using Self-Consistency}
Consistency across stochastic responses has been proven to be a strong signal for improving reasoning and code generation \citep{chen2024universal, wang2023selfconsistency}. Building on this, SelfCheckGPT \citep{manakul2023selfcheckgpt} uses agreement among diverse model outputs as an indicator of hallucination. HaLo \citep{elaraby2023halo} used consistency-based metrics to detect sentence-level hallucinations in the generations. Atomic Self-Consistency (ASC) \citep{thirukovalluru2024atomicASC} extends consistency-based methods by decomposing multiple stochastic responses into atomic facts, clustering them to reduce redundancy, and using cluster strength as a proxy for factual consistency. Inspired by ASC, we leverage atomic-level consistency signals to construct preference pairs for alignment.

\subsection{Alignment Methods}\label{ss:alinment_methods}
Although inference-time methods have proven effective in reducing hallucinations, they are often computationally expensive, typically relying on multiple stochastic generations or repeated LLM queries to verify individual atomic facts within a response. To address this, recent work has focused on alignment-based training approaches that aim to induce factuality during training thus reducing the inference-time costs.


FactAlign \citep{huang2024factalign}, a strong supervised baseline, leverages Kahneman-Tversky Optimization (KTO) to align models using atomic fact labels from FactScore, which identifies individual facts using GPT-3.5 models and verifies them via a Wiki-based retriever.

SKT \citep{zhang2024self} uses GPT-3.5-models to first generate atomic facts and then verifying questions from multiple stochastic responses. An external retriever is then used to score each atomic fact, with scores aggregated to produce response-level ratings. Similarly, FactTune \citep{tian2023fine} generates atomic claims and corresponding questions using GPT-3.5, and scores them with an external retriever, aggregating claim-level scores to obtain response-level scores. In both methods, these scores are used to construct preference pairs—high-scoring responses as preferred and low-scoring as non-preferred. These pairs are then used for DPO-based alignment training of the model.


SKT and FactTune also propose self-supervised variants of their methods, wherein the base model is directly used to score atomic factuality instead of relying on an external retriever. However, these variants are \textit{not truly self-supervised}, as they still depend on an additional GPT-3.5-based pipeline to generate atomic claims and verification questions. Assuming $m$ stochastic responses are generated, both methods require $m$ base LLM calls for the initial generations. This is followed by approximately $(m\times f\times2)$ GPT-3.5 calls for generating $f$ atomic claims and verification questions per response. Finally, the self-evaluation scores are computed using an additional $(k\times m\times f)$ base LLM calls, where $k\approx1$ for SKT and $k\approx20$ for FactTune—making the overall process extremely expensive. FactTune \cite{tian2023fine} also offers a GPT-3.5–free variant using entity recognizers for fact extraction, but it performs notably worse than the GPT-3.5-based version.

In terms of scoring, SKT leverages the generated atomic claims to estimate self-evaluation scores, while FactTune disregards the claims entirely and bases its confidence estimation solely on the calibration of atomic questions to estimate a response confidence. Notably, for factual calibration, self-consistency-based approaches have been shown to outperform self-evaluation-based scoring—as used in SKT—and other methods like atomic question confidence, as used in FactTune \citep{huang2024calibrating}. Recent work shows that self-consistency–driven preference tuning significantly outperforms other baselines on reasoning tasks \citep{prasad2024self}.

Motivated by these findings, we propose Atomic Consistency Preference Optimization (\acpo), which leverages atomic self‑consistency—the agreement of individual facts across stochastic responses—to score outputs and construct preference data for DPO‑based alignment. \acpo\ generates 
$m$ stochastic responses using only $m$ base‑LLM calls and eliminates the need for costly atomic fact labeling or large‑model verification (e.g., GPT‑3.5). Instead, it employs a lightweight embedding model to cluster atomic facts and uses cluster strengths as a measure of consistency. This design substantially improves efficiency while fostering strong factual consistency. Section \ref{sec:dpo} discusses basics of DPO followed by our methodology in Section \ref{sec:method}.
\acpo\ generates $m$ stochastic responses using $m$ base LLM calls and eliminates the need for costly atomic fact annotation or large model verification.


\section{Background: DPO Alignment Mechanism}\label{sec:dpo}
Reinforcement Learning with Human Feedback (RLHF) has become a foundational approach for aligning large language models (LLMs) with human preferences and reducing hallucinations \citep{tian2023fine,zhang2024self}. This line of work began with InstructGPT, which introduced a reward model and Proximal Policy Optimization (PPO) for fine-tuning \citep{ouyang2022training}. To reduce the cost of human annotations and leverage the growing capabilities of LLMs, later approaches such as Constitutional AI \citep{bai2022constitutional} and RLAIF \citep{Lee2024} replaced human preferences with model-generated critiques. More recently, Direct Preference Optimization (DPO) \citep{rafailov2023direct} simplified this process by eliminating the need for a separate reward model and complex reinforcement learning, instead directly optimizing log-likelihood ratios over preference pairs. In this work, we create self-supervised preference data and adopt DPO to perform alignment tuning.



We apply the standard DPO loss function, shown in Equation~\ref{eq:DPOloss}.
\begin{equation}
\label{eq:DPOloss}
\begin{aligned}
&\mathcal{L}_{\text{DPO}}(\pi_\theta; \pi_{\text{ref}})
= - \mathbb{E}_{(x, y_w, y_l) \sim \mathcal{D}}  
\Bigg[\\
&\log \sigma \Big( 
\beta \log \tfrac{\pi_\theta(y_w \mid x)}{\pi_{\text{ref}}(y_w \mid x)}  
\Big)
- \log \sigma \Big( 
\beta \log \tfrac{\pi_\theta(y_l \mid x)}{\pi_{\text{ref}}(y_l \mid x)} 
\Big) 
\Bigg]
\end{aligned}
\end{equation}

 The DPO approach fine-tunes a policy $\pi_\theta$ by maximizing the preference margin between a preferred response $y_w$ and a less preferred one $y_l$, relative to a reference policy $\pi_{\text{ref}}$. The $\beta$ controls how aggressively the model separates preferred from non-preferred responses.

\begin{figure*}[t]
    \centering
    \includegraphics[width=0.97\textwidth]{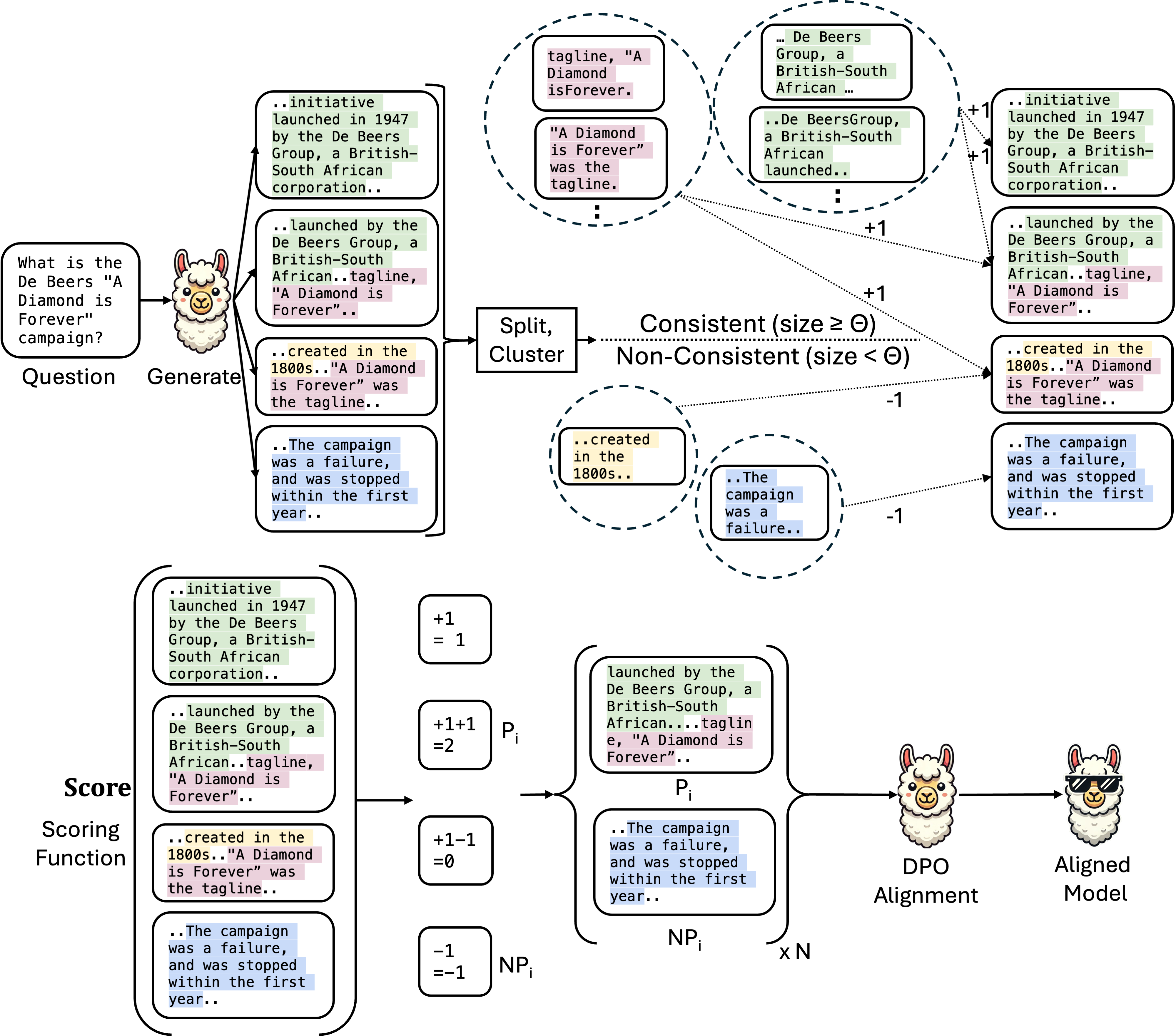}
    \caption{\acpo\ data curation pipeline. Steps 1–5 (Top): Generate stochastic responses for a question, Split and extract atomic facts; Cluster the atomic facts; Identify consistent and non-consistent clusters; Score responses based on cluster consistency. Step 6-7 (Bottom): Highest and lowest scoring curated as preference pairs; DPO alignment.
    }
    \label{fig:main}
    \vspace{-1.3em}
\end{figure*}

\section{Methodology}\label{sec:method}
In this section, we present our \acpo\ framework, detailing the training data generation process and the fine-tuning methodology.

\subsection{Overview}
We leverage the model's generation capabilities to produce $m$ stochastic responses for a given prompt $P$. Following the ASC~\citep{thirukovalluru2024atomicASC} framework, each response $R_i$ is decomposed into a set of atomic facts $[a_1,a_2 \dots a_k]$. All atomic facts from all responses are then aggregated and clustered. The core idea from ASC is that atomic facts appearing in larger clusters are more likely to be factual; we refer to these as consistent clusters $\mathcal{C}_i$, while smaller clusters form Non-consistent clusters $\mathcal{NC}_i$. For each atomic fact in a response $R_i$, we determine whether it belongs to a consistent or non-consistent cluster. If the atomic fact belongs to a consistent cluster, we reward its initial response (by adding a positive score); if not, we apply a penalty. This results in a consistency-based score for each response $R_i$, allowing us to distinguish between preferred and non-preferred responses. Next, we describe the scoring mechanism and training data generation for DPO alignment in detail.

\subsection{Data Generation}\label{sec:Data_create}
\subsubsection{Step 1: Initial Responses Generation}
Given a question $q$, our objective is to prompt a large language model $L$ to generate a response that is both accurate and informative. To achieve this, we adapt the system prompt from FactTune~\citep{tian2023fine}, modifying it to:
\textit{``You are an intelligent assistant who answers questions accurately''}. This modified prompt is then concatenated with the input of the actual question $q$. As a result, we obtain $m$ independent responses denoted as $[R_1, R_2, \ldots, R_i, \ldots, R_m]$ by querying the model $L$ with $q$ using the predefined prompt $P$.

\subsubsection{Step 2: Splitting Initial Responses for Atomic Facts}
We then decompose each candidate response into a set of atomic statements. A single response $R_i$ to a question may contain multiple sentences, each potentially expressing one or more atomic facts. While prior work~\citep{min2023factscore,zhang2024self,huang2024factalign} has employed large instruction-tuned models, like GPT-3.5, to identify atomic facts from long-form text, these methods are often computationally expensive and lack scalability. Inspired by~\citet {arslan2020benchmark,thirukovalluru2024atomicASC,liu2023evaluating}, we adopt a simplified yet effective alternative: treating each sentence in a generation as an atomic fact. Specifically, following the ASC paper, we apply standard sentence tokenization techniques~\citep{bird2009natural} to segment each response into individual sentences, which we regard as atomic facts. After tokenization, the $i$-th response $R_i$ is represented as a list of atomic sentences (atomic facts) $[a_{i1}, a_{i2}, \ldots, a_{ik}]$, where k is the atomic fact count. Note that although ACPO uses sentences as atomic facts, it is compatible with any atomic fact identification method (e.g., GPT-4), after which the subsequent steps can be applied.

\subsubsection{Step 3: Clustering Atomic Facts}
To address the high computational cost of verifying the relevance of each atomic fact across multiple generated responses, we follow the ASC framework by clustering semantically similar atomic units. ASC applies agglomerative clustering on sentence embeddings obtained from SimCSE~\citep{gao2021simcse} (a lightweight BERT-based sentence embedder), leveraging the substantial semantic overlap across generations to group atomic facts with similar meanings. Although agglomerative clustering has cubic worst-case complexity, it remains substantially more efficient than knowledge-base or LLM-based verification for each atomic fact.


\subsubsection{Step 4: Consistent (\texorpdfstring{$\mathcal{C}$}{C}) and Non-consistent (\texorpdfstring{$\mathcal{NC}$}{NC}) Clusters}

Our method leverages the inherent consistency of model outputs that are quantified by the size of each cluster. Clusters with count below a threshold $\Theta$ are determined as $\mathcal{NC}_i$, while those above or equal to the $\Theta$ are classified as $\mathcal{C}_i$. The hypothesis is that LLMs are knowledgeable, and the high-frequency information in responses is more factual than rare ones \citep{wang2024fine}. Therefore, information in $\mathcal{C}_i$ is more factual, and we utilize this property to score the initial responses $R$. 

\subsubsection{Step 5: Scoring Function}
We define a consistency-based scoring function for each response $R_i$ based on the classification of its atomic facts into consistent and non-consistent clusters. Let the atomic facts extracted from response $R_i$ be:  
$R_i = [a_{i1}, a_{i2}, \ldots, a_{ik}]$.
Let $\mathcal{C}_i$ denote the set of consistent clusters (with size $\geq \Theta$), and $\mathcal{NC}_i$ denote the set of non-consistent clusters (with size $< \Theta$). We score each response $R_i$ as:
\begin{equation}
\begin{aligned}
\label{eq: Score}
\text{Score}(R_i) &= \sum_{j=1}^{k} \delta(a_{ij});
\quad\text{where} 
\quad\\
\delta(a_{ij}) &=
\begin{cases}
+1, & \text{if } a_{ij} \in \mathcal{C}_i \\
-1, & \text{if } a_{ij} \in \mathcal{NC}_i \\
0,  & \text{otherwise}
\end{cases}
\end{aligned}
\end{equation}

This scoring mechanism rewards atomic facts belonging to consistent clusters and penalizes those from non-consistent clusters.  
For example, if $R_i = [a_{i1}, a_{i2}, a_{i3}]$, and $a_{i1}, a_{i3} \in \mathcal{C}_i$ while $a_{i2} \in \mathcal{NC}_i$, then:  
$\text{Score}(R_i) = 1 + (-1) + 1 = 1$.

\subsubsection{Step 6: Preference Data Obtain}
After Steps 1-5, each $R_i$ is assigned a consistency-based score. To construct preference pairs, we sort all responses by their scores and select the top-$1$ response as preferred ($P_i$) and the bottom-$1$ response as non-preferred ($NP_i$). This results in a training dataset:
$
\mathcal{D} = \{(x_i, P_i, NP_i)\}
$, containing $|\mathcal{D}|$ (dataset size) datapoints where $x_i$ is the prompt.

\subsection{Step 7: DPO Alignment}
The preference data pairs generated in \S~\ref{sec:Data_create} are subsequently used for DPO alignment, with the detailed training setup described in \S~\ref{sec:Training_Detail}.

\section{Experiments}
 \begin{table*}[t]
 \centering
 \resizebox{0.92\textwidth}{!}{
 \begin{tabular}{l||cc|cc||cc|cc}
 \toprule
                             & \multicolumn{4}{c||}{\textbf{\llamat}}                                                                                                        & \multicolumn{4}{c}{\textbf{\phit}}                                                                                                                          \\
                             \cline{2-9}
                             & \multicolumn{2}{c|}{\textbf{LongFact}}                                        & \multicolumn{2}{c||}{\textbf{BioGen}}                              & \multicolumn{2}{c|}{\textbf{LongFact}}                                        & \multicolumn{2}{c}{\textbf{BioGen}}                                            \\
\textbf{Method}             & \multicolumn{1}{l}{\textbf{Score}} & \textbf{\#Claim} & \textbf{Score}       & \textbf{\#Claim} & \textbf{Score} & \textbf{\#Claim} & \textbf{Score} & \textbf{\#Claim} \\\midrule
\rawmodel                    & 79.8                               & 121.2                                   &  55.9 & 61                    & 78                                 & 90.2                                    & 41.7                               & 88.4                                    \\
\factalign                  & \textbf{83.3}                               & 119.9                                   & 57.1                 & 56.7                                    & 81.2                               & 113.8                                   & 47.1                               & 100.9                                   \\
\acpo\ & 82.1                               & 143.8                                   & \textbf{58}                   & 68                                      & \textbf{84.6}                               & 70.7                                    & \textbf{51.8}                               & 67.1                                    \\
\bottomrule
\end{tabular}}
\caption{Factscore accuracy for \acpo\, \factalign\ and other baselines on LongFact, Bios datasets. \factalign\ and \acpo\ were trained on the train set of LongFact. \#Claim is the average number of claims produced by the model.}\label{tab:1}
\vspace{-0.5em}
\end{table*}
\subsection{Models and Baselines}
We conduct a comprehensive comparison of our proposed method, \ours, against two key baselines. The first is \factalign\ \citep{huang2024factalign}, a recently introduced alignment technique that leverages fine-grained, atomic fact-level annotations provided by the FactScore benchmark to guide the alignment process. The second baseline is the unaligned model, \rawmodel, which serves as a reference point to assess the impact of alignment strategies. To ensure a fair and thorough evaluation, we perform experiments at two different model scales: Phi-3 Mini (4B) and Llama3 (8B), allowing us to assess performance across varying levels of model capacity. \acpo\ uses $m=30$ following \citet{zhang2024self}. $\beta$ is set to 0.1 during \acpo\ training. More details in \S \ref{sec:Training_Detail}. We are unable to compare with SKT due to unavailable code, and with FactTune as it requires ~2M GPT-3.5 calls on LongFact dataset and uses private datasets during their training, preventing cross-evaluation.
\subsection{Datasets and Evaluation}

The training split of the LongFact dataset \citep{wei2024long}, consisting of 2,097 examples, was used to align both \factalign\ and the proposed model, \acpo. Evaluation was conducted on the test splits of the LongFact and BioGen datasets \citep{min2023factscore}, with results reported for all models and baselines. Both LongFact and BioGen are English-language datasets. As FactScore relies on topic names—which are not available for LongFact—GTR-XL was used to retrieve the most relevant documents from the full Wikipedia corpus to support factual grounding during evaluation. FactScore reports two key metrics: the factual precision of the claims present in the output and the total number of factual claims identified in the output. The former is the more important metric.

\subsection{Main Results}
Table~\ref{tab:1} shows the comparison between \acpo\ and other methods. Despite not relying on any external signals like \factalign, \acpo\ outperforms it in three of the four settings, achieving an average gain of $+1.95$ points on Phi-3 and Llama3 across the LongFact and BioGen benchmarks. With Llama3‑8B, \acpo\ also produces outputs containing a greater number of claims. This stems from its use of the ASC principle to identify preferred and non‑preferred responses for training. As noted by
\citet{huang2024calibrating}, models vary in their calibration
 (i.e., consistency across stochastic responses),
which we believe explains why Phi-3 generates
fewer claims.

\subsubsection{Which Models Benefit Most from ACPO?}\label{ap:model_calibration}

We used a very small threshold ($\Theta$=2) for identifying consistent clusters for both Phi-3 and Llama3. Note that \acpo\ leverages the confidence of a model in generated responses to pick preferred/non-preferred data. Not all models are equally calibrated (confident about their responses). Some models are more calibrated than others \cite{huang2024calibrating}. Hence, some models might perform better with \acpo\ than others.

Table \ref{tab:consistency_ratio} shows the ratio of the number of consistent clusters and non-consistent ones. Despite the same small threshold ($\Theta$), we note that Phi-3 has a much higher number of non-consistent clusters. Fewer consistent clusters suggest that the model is less calibrated compared to Llama3. Hence, Phi-3 tends to pick smaller responses as preferred ones (this can avoid the negative score from non-consistent clusters). As shown in Table \ref{tab:5}, one can relax the constraint from \acpo\ to balance the length of the training dataset.

\begin{table}[h]
\centering
\resizebox{\columnwidth}{!}{
\begin{tabular}{l c}
\toprule
\textbf{Training Data} & \makecell{\textbf{(\#Consistent:}\\\textbf{\#Non-Consistent})} \\
\midrule
\phit   & (1:3) \\
\llamat & (1:1.8) \\
\bottomrule
\end{tabular}}
\caption{Ratio of consistent to non-consistent data across training sets.}
\label{tab:consistency_ratio}
\vspace{-1em}
\end{table}



\begin{table*}[t]
\centering
\resizebox{\textwidth}{!}{
\renewcommand{\arraystretch}{1.2}
\begin{tabular}{l||cc|cc|cc||cc|cc|cc}

\toprule
                            & \multicolumn{6}{c||}{\textbf{LLama-3-8b-Instruct}}      & \multicolumn{6}{c}{\textbf{Phi-3-mini-4k-instruct}}\\\cline{2-13}
                            & \multicolumn{2}{c|}{\textbf{Length}}                              & \multicolumn{2}{c|}{\textbf{LongFact}                  }& \multicolumn{2}{c||}{\textbf{BioGen}}                            & \multicolumn{2}{c|}{\textbf{Length}}                              & \multicolumn{2}{c|}{\textbf{LongFact}                  }& \multicolumn{2}{c}{\textbf{BioGen}}                               \\\cline{2-13}
\textbf{Method}             & \multicolumn{1}{l}{\textbf{P}} & \textbf{NP} & \textbf{Score} & \textbf{\#C} & \textbf{Score} & \textbf{\#C} & \textbf{P} & \textbf{NP} & \textbf{Score} & \textbf{\#C} & \textbf{Score} & \textbf{\#C} \\\midrule
\acpo\ & 478                            & 457                             & \textbf{82.1}                               & 143.8                                   & 58             & 68                                      & 307                            & 327                             & 84.6                               & 70.7                                    & \textbf{51.8}                               & 67.1                                    \\
\acpo\ (5,5)      & 466                            & 449                             & 79.6                               & 150.6                                   & 57.3           & 83.2                                    & 287                            & 295                             & 85.1                               & 72.5                                    & 49.4                               & 64.3                                    \\
\acpo\ (5,4+1)  & 466                            & 461                             & 80.2                               & 140.2                                   & 58.2           & 73.9                                    & 287                            & 285                             & \textbf{85.4}                               & 75.6                                    & 50.5                               & 65.4                                    \\
\acpo\ (5,3+2) & 466                            & 468                             & 81.4                               & 127.8                                   & \textbf{59.7}           & 68                                      & 287                            & 279                             & 84.9                               & 75.2                                    & 50.6                               & 67.1\\
\bottomrule
\end{tabular}}
\caption{Using length as an additional criterion to balance preferred and non-preferred data leads improves \#Claims, thereby increasing recall. Results are better for Llama3 on BioGen and for Phi-3 on LongFact.}\label{tab:5}
\vspace{-1em}
\end{table*}
\begin{table}[ht]
\centering
\begin{tabular}{lcc}
\toprule
\textbf{Method} & \textbf{F1} & \textbf{\#Claim} \\
\midrule
\rawmodel & 75.88 & 100.85 \\
\factalign\ & 80.32 & 103.33 \\
\textbf{\acpo\ (Ours)} & \textbf{83.84} & \textbf{120.17} \\
\bottomrule
\end{tabular}
\caption{Longfact F1 score with Llama-3-8b-Instruct}
\label{tab:longfact-results}
\vspace{-1em}
\end{table}

\begin{table}[ht]
\centering
\resizebox{\columnwidth}{!}{
\renewcommand{\arraystretch}{1.1}
\begin{tabular}{l|c|c|cc}
\toprule
                   & \multicolumn{4}{c}{\textbf{Phi-3-mini-4k-instruct}}\\
                   & \multicolumn{1}{l}{\textbf{}}       & &\multicolumn{2}{c}{\textbf{BioGen}}                                            \\\cline{2-5}
\textbf{Model}     & \multicolumn{1}{l|}{\textbf{Inf.}} & \textbf{$\text{Score}(R_i)$} & \multicolumn{1}{l}{\textbf{Score}} & \multicolumn{1}{l}{\textbf{\#Claims}} \\\midrule
\multirow{2}{*}{\rawmodel}           & Direct                              & -5.61 & 42.9                               & 85.3                                    \\
          & ASC                                 & 4.10 & 45.7                               & 84.5                                    \\\hline
\multirow{2}{*}{\acpo\ (It. 1)} & Direct                              & 1.27 & \underline{51.8}                               & 67.1                                    \\
 & ASC       & 7.97 & \textbf{54}                                 & 70.7                                    \\\hline
\multirow{2}{*}{\acpo\ (It. 2)} & Direct                              & 2.30 & 48.2                                 & 80.5                                       \\
& ASC                                 & 10.83 & \underline{51.8}              & 93.4\\
\bottomrule
\end{tabular}}
\caption{Performance with ASC at inference time. \acpo\ with direct inference outperforms \rawmodel, showing the importance of ASC in training. Applying ASC at inference further boosts performance, with Iteration 1 performing best. Later iterations add no gains but still beat \rawmodel\ while generating more \#Claims, highlighting ASC’s value for alignment data.}\label{tab:3}
\vspace{-1.5em}
\end{table}

\begin{table}[h] \centering \resizebox{\columnwidth}{!}{ \begin{tabular}{l|c|c|c} \toprule \textbf{\diagbox{Stages}{Methods}} & \textbf{\acpo} & \textbf{\makecell{Sentence\\+GPT-4}} & \textbf{\makecell{GPT-4\\(SKT)}} \\ \midrule \texttt{Embedding} & 2.9 s & - & - \\ \texttt{Clustering} & 0.39 s & - & - \\ \texttt{Generating Atomic Facts} & - & - & 594.9 s \\ \texttt{Verifying Atomic Facts} & - & 922.1 s & 781.1 s \\\hline \textbf{Total} & \textbf{3.3 s} & 922.1 s & 1376.0 s \\ \bottomrule \end{tabular}} \caption{Comparison of computational time across different methods for preference pair construction. Our \acpo\ employs a clustering-based approach, whereas Sentence+GPT-4 treats sentences as atomic facts and verifies them using GPT-4, and GPT-4 (SKT) both generates and verifies atomic facts through GPT-4.} \label{tab:runtime} \end{table}


\begin{table}[t]
\centering
\resizebox{0.94\columnwidth}{!}{
\begin{tabular}{l|cc}
\toprule
& \multicolumn{2}{c}{\textbf{BioGen}}\\
\textbf{Method} & \textbf{Score} & \textbf{\#Claim} \\\midrule
\acpo                           & 51.8 & 67.1 \\
\acpo\ \texttt{(5,5)} & 49.4 & 64.3 \\
\acpo\ \texttt{w/o NC Penalty} & 46.9 & 99.7 \\
\hline
\texttt{Longest Preferred}       & 41.3 & 97.6 \\
\texttt{Shortest Preferred}      & 42.8 & 10.6 \\
\bottomrule
\end{tabular}}
\caption{Stronger preference signals in \acpo\ perform better than weaker ones in \acpofive. Not penalizing non-consistent atomic facts yields worse alignment. Favoring short or long responses harms factual precision, highlighting the value of \acpo’s alignment strategy.}\label{tab:4}
\vspace{-1.5em}
\end{table}
\begin{table}[h]
\centering
\begin{tabular}{lcc}
\toprule
\textbf{Method} & \textbf{Score} & \textbf{\#Claims} \\
\midrule
\acpo ($\Theta=2$)            & 51.8 & 67.1 \\
\acpo ($\Theta=3$)            & 50.4 & 48.7 \\
\bottomrule
\end{tabular}
\caption{ACPO performance under different $\Theta$ values.}\label{tab:theta}
\vspace{-1em}
\end{table}

\begin{table}[ht]
\centering
\resizebox{\columnwidth}{!}{
\begin{tabular}{l|cccccc|c}
\toprule
\textbf{Temp.} & 0 & 0.3 & 0.6 & 0.9 & 1.2 & 1.5 & \acpo\ \\\midrule
\textbf{Score}       & 42.9 & 43.2 & 41.7 & 38.4 & 36.3 & 33.3 & \textbf{51.8} \\
\textbf{\#Claim}     & 85.3 & 85.2 & 88.4 & 93   & 99.9 & 111 & 67.1 \\
\bottomrule
\end{tabular}}
\caption{Self-supervised alignment in \acpo\ (temp=0) outperforms temperature scaling—a simple post-hoc calibration method—across multiple temperatures.}
\label{tab:2}
\vspace{-1em}
\end{table}

\begin{table*}[ht]
\centering
\scriptsize
\begin{tabular}{>{\raggedright\arraybackslash}p{0.48\textwidth}|>{\raggedright\arraybackslash}p{0.48\textwidth}}
\toprule
\multicolumn{2}{c}{\textbf{Question} | "What is the De Beers diamond monopoly controversy?}\\\midrule
\textbf{Preferred} & \textbf{Non-Preferred}\\\hline
\hlgreen{The De Beers diamond monopoly controversy centers ... by the De Beers company, ... had a near-complete monopoly over the global diamond market.} \hlgreen{Several key details and examples include: 1. De Beers' monopoly began in the late 19th century when Cecil Rhodes, a British businessman and mining magnate, founded the company in South Africa in 1888.} \hlgreen{2. By the early 20th century, De Beers ... and controlled over 90\% of the world's rough diamond production and trade.} \hlgreen{3. The company's ... stockpiling diamonds and controlling ... to maintain high prices and ensure a stable supply.} ... \hlgreen{7. In 2005, the Antitrust Division of the United States Department of Justice filed an antitrust lawsuit ... price-fixing and maintaining a monopoly in the diamond market.}... & \hlgreen{The De Beers diamond monopoly controversy revolves around the company's control over the global diamond market in the ... early 21st centuries}. \hlgreen{De Beers, a British company founded in 1888 by Cecil Rhodes, came to dominate the diamond industry by controlling diamond production and distribution.} ... \hlpink{De Beers' monopoly was challenged by a South African-based company, Cut Stone, led by Ernest Oppenheimer, grandson of ..., Cecil Rhodes}. \hlpink{Cut Stone accused De Beers of fixing diamond prices and controlling the market.} \hlpink{The controversy intensified when the South African government, ..., announced plans to nationalize the diamond industry in 1981.} ... \hlpink{In 1998, a class action lawsuit was filed against De Beers in the United States by the American Antitrust Institute (AAI)}.... \\\bottomrule
\end{tabular}
\caption{Preferred and Non-Preferred responses curated by \acpo. Green highlights indicate phrases verified to be correct, while red highlights mark incorrect phrases according to Wikipedia. In this example, terms like Cutstone, the nationalization claim, AAI are all hallucinated.}
\label{tab:qualitative_fulltext}
\vspace{-1.5em}
\end{table*}

\subsection{Analysis 1: Can Length Balancing Alignment data help?}
Our analysis of the alignment training data revealed distinct trends in preference behavior across the Phi-3 and Llama3 datasets. Specifically, in Phi-3, preferred responses were generally shorter than non-preferred ones, whereas in Llama3, the opposite trend was observed—preferred responses tended to be longer. To evaluate whether explicitly incorporating response length into the training signal could enhance the performance of \acpo, we explored several variants that adjusted the length of non-preferred responses in alignment with these trends—shortening them for Phi-3 and lengthening them for Llama3.

Specifically, \acpo\ selects the single highest- and lowest-scoring responses as the preferred and non‑preferred examples, respectively. In contrast, \acpo~\texttt{(5,5)} expands this selection to the top five and bottom five responses while keeping the number of training steps unchanged. Building on this variant, we introduced a length‑based modification, replacing one or two of the non‑preferred responses with alternatives chosen by length—favoring longer responses for Phi‑3 and shorter ones for Llama3. As shown in Table~\ref{tab:5}, this adjustment yields performance improvements. Additionally, length balancing increases the \#Claims, which is valuable in scenarios prioritizing high recall. This is because \acpo’s selection process does not constrain response length, while the factual precision metric is length‑sensitive—short responses can inflate precision scores \citep{huang2024factalign}.

\subsection{Analysis 2: Measuring Recall}
While Table \ref{tab:1} reports results for factual precision, recall is also critical in certain scenarios. Therefore, we additionally compute the F1 score of different models under LongFact using their custom API. It is important to note that FactScore does not provide an F1 metric; hence, we do not report it. As shown in Table~\ref{tab:longfact-results}, \acpo\ outperforms other methods.

\subsection{Analysis 3: \acpo’s Efficiency in Preference Pair Construction}
Compared to LLM-based alignment approaches, \acpo\ achieves a substantial improvement in computational efficiency. As reported in Table~\ref{tab:runtime}, \acpo\ processes and constructs preference data in just $3.3$ seconds per example ($2.9$ s for embedding and $0.39$ s for clustering), whereas \textit{Using Sentences as Atomic Facts and Verifying Them Using GPT-4} requires $922.1$ s, and \textit{Generating Atomic Facts Using GPT-4 and Verifying Them Using GPT-4 (SKT)} requires $1376.0$ s. This corresponds to an efficiency gain exceeding two orders of magnitude, primarily due to \acpo's fully self-supervised design, which eliminates the need for factual verification via large language models.

\subsection{Ablation 1: \acpo\ compared with Inference time ASC}
This ablation study investigates whether explicit alignment training is necessary or if the ASC principle can be effectively applied directly at inference time to achieve performance comparable to or better than \acpo. Although applying ASC at inference time is computationally more expensive, we assess its effectiveness relative to aligned models. Results are shown in Table~\ref{tab:3}. Direct refers to generating responses from the raw or trained model without any additional sampling, whereas ASC generation involves sampling multiple stochastic outputs and selecting the highest-scoring one based on $\text{Score}(R_i)$. As observed, \acpo\ with direct decoding outperforms ASC applied to the unaligned \rawmodel, suggesting that incorporating ASC into the training process to construct a preference dataset leads to more substantial improvements than applying it only at inference time. Moreover, applying the ASC on top of \acpo\ yields further gains. Motivated by the improvements from \acpo+ASC, we conducted an additional round of self-supervised training using the already-aligned model. But this attempt did not improve test performance—likely due to overfitting after 25 epochs of \acpo\ training (\S\ref{sec:Training_Detail}). Iteration 2 had a higher $\text{Score}(R_i)$ score than Iteration 1, which suggests that, although the model became more internally consistent, the improvement did not generalize to the test set, likely due to overfitting on the LongFact training data.


\subsection{Ablation 2: Dissecting \acpo\ Scoring Mechanism}
To understand the contribution of individual components within \acpo, we conduct a series of studies, with results summarized in Table~\ref{tab:4}. In its full form, \acpo\ rewards responses that include atomic facts from consistent clusters and penalizes those containing atomic facts from non-consistent ones.

We examine \acpo~\texttt{(5,5)} alongside \acpo~\texttt{w/o Non‑Consistent Penalty}, which removes the penalty for selecting atomic facts from non‑consistent clusters. To probe the effect of response length on alignment, we further extend our ablation by explicitly preferring either the longest or shortest responses, using a randomly selected response as the negative in both cases. Without the penalty, the model tends to favor longer responses, often ranking them more highly. The similarity in the number of factual claims between the penalty‑free variant and \texttt{Longest Preferred} suggests that removing the penalty encourages behavior akin to explicitly favoring longer responses, underscoring the penalty’s role in guiding alignment toward more precise and reliable outputs. Notably, the \texttt{Shortest Preferred} variant produces very few factual claims and performs poorly on FactScore precision, indicating that simply favoring brevity is not an effective strategy for improving factual alignment.

\subsection{Ablation 3: Effect of \texorpdfstring{$\Theta$}{Theta}}
A higher value of $\Theta$ encourages the model to favor highly consistent responses, which often leads to the selection of shorter responses, as maintaining consistency is easier with fewer facts. Results are shown in Table~\ref{tab:theta}.

\subsection{Analysis 4: Can simple calibration techniques match \acpo\ performance?}
\acpo\ is a simple self-supervised algorithm that uses the atomic consistency principle to align the model to generate better responses. Temperature scaling is another way of model calibration \citep{renze2024effect}. We investigate if the gains in \acpo\ can be achieved by simple temperature scaling of the \rawmodel. Table~\ref{tab:2} shows the results. \acpo\ significantly outperforms all temperature settings of the \rawmodel.

\subsection{Analysis 5: Qualitative Analysis}\label{ss:qa}
Preferred and non-preferred examples curated by \acpo\ are shown in Table \ref{tab:qualitative_fulltext}. As the highlights indicate, \acpo\ identifies high-quality examples without relying on external signals.

\section{Conclusion}

We introduce Atomic Consistency Preference Optimization (\acpo), a self-supervised method for aligning LLMs to improve factual accuracy in long-form question answering. \acpo\ leverages the atomic self-consistency principle to curate high-quality preference data pairs which when used alignment eliminates the need for external supervision or stronger LLMs. Extensive experiments show that \acpo\ outperforms a strong supervised baseline (FactAlign) on LongFact and BioGen benchmarks. We further show that using atomic consistency during training is more beneficial and leads to better factual precision than solely applying it at inference time. Additional ablations validate the length penalties and the robustness of \acpo\ across different model sizes. In summary, \acpo\ presents a simple, effective, and efficient self-supervised approach to enhance factual alignment in LLMs, leading to more trustworthy and factual generation capabilities.

\section{Limitations}
The self consistency principle can be combined with other self evaluation methods to get hybrid alignment frameworks that combine the strengths of both paradigms. However, in this study, we focused on studying self-consistency based methods for factuality.

\section{Ethics Statement}
While our model is not tied to any specific applications, it could be used in
sensitive contexts such as health-care, etc.  Any work using our
method is requested to undertake extensive quality-assurance and
robustness testing before applying in their setting. To the best of our knowledge, datasets used in our work do not contain any sensitive information. 

\section{Reproducibility Statement}
Code: \href{https://github.com/JingfengSteven/ACPO}{https://github.com/JingfengSteven/ACPO}

\noindent\textbf{License}: Datasets and methods utilized in this study are under the Apache License 2.0 or the MIT License. This research adheres to the respective licensing terms. Outputs of this work are released under the Apache License 2.0.\\

\newpage

\bibliography{custom}
\newpage
\appendix
\section{Appendix}

\begin{table*}[ht]
\centering
\resizebox{\textwidth}{!}{
\begin{tabular}{l||cc|cc||cc|cc}
\toprule
                            & \multicolumn{4}{c||}{\textbf{\llamat}}                                                                                                        & \multicolumn{4}{c}{\textbf{\phit}}                                                                                                                          \\
                            \cline{2-9}
                            & \multicolumn{2}{c|}{\textbf{LongFact}}                                        & \multicolumn{2}{c||}{\textbf{BioGen}}                              & \multicolumn{2}{c|}{\textbf{LongFact}}                                        & \multicolumn{2}{c}{\textbf{BioGen}}                                            \\
\textbf{Method}             & \multicolumn{1}{l}{\textbf{Score}} & \textbf{\#Claim} & \textbf{Score}       & \textbf{\#Claim} & \textbf{Score} & \textbf{\#Claim} & \textbf{Score} & \textbf{\#Claim} \\\midrule
\rawmodel                               & 79.5 & 121.6 & 55.3 & 61.1 & 79.8 & 91.1 & 42.9 & 85.3 \\
\factalign                             & \textbf{83.1}                 & 118.2                & 57.6 & 58.1 & 82.6                 & 112.2                & 48.4 & 97.3 \\
\acpo & 82.1                 & 143.8                & \textbf{58}   & 68   & \textbf{84.6}                 & 70.7                 & \textbf{51.8} & 67.1\\
\bottomrule
\end{tabular}}
\caption{\acpo\ vs other methods at temperature=0 during inference}\label{tab:ap1}
\end{table*}

\subsection{Training Details}\label{sec:Training_Detail}
The alignment procedure is adapted from \citet{tian2023fine}. For training, we set the batch size to 32 for the Phi-3-mini-4k-instruct model and 64 for the LLaMA-3-8B-Instruct model. We use a linear warmup learning rate schedule, with 100 warmup steps for the LLaMA model and 150 for the Phi model, followed by cosine decay. The learning rate is kept as default=$1e^{-6}$, and $\beta$ is set as 0.1. Rather than using a fixed number of epochs, training is controlled by the total number of steps. Given that we use either \acpo\ (5,5) $(5\times5=25)$ preference pairs per question or \acpo\ (1,1) $(1\times1=1)$ preference pairs per question, we ensure 1 complete epoch for the 25-pair case. Consequently, for the 1-pair case, we train for 25 epochs to maintain step parity. Gradient clipping is applied with a default threshold of 10. The total training time is approximately 1 hour for the Phi model and 2.5 hours for the LLaMA model, using 4 NVIDIA H800 GPUs with 80 GB of memory each.

We used the default temperature values - 0.5 for \factalign\ \citep{huang2024factalign} and 0.6 for \rawmodel. For \acpo, we use greedy decoding (temperature = 0) to ensure reproducibility and to evaluate the model's capability without introducing randomness. Results for greedy decoding of all models present in \S \ref{tab:ap1}. \acpo\ beats \factalign\ even in this setup.

\subsection{Clustering Details}
We employ Agglomerative Clustering with average linkage and cosine distance as the similarity metric. The number of clusters is determined dynamically by setting $n\_clusters=None$ and applying a $distance\_threshold=0.15$, such that clusters are continuously merged until the pairwise inter-cluster distance exceeds the specified threshold. The $\Theta$ value for consistent and non-consistent filtering is set as 2.

\subsection{Results with Greedy (temperature=0)}
Table~\ref{tab:ap1} shows the results for all models at temperature=0. \acpo\ beats baselines and \factalign\ even at greedy decoding (temperature=0).

\subsection{Data Sheet}
 We present the dataset details along with key statistics relevant to the clustering process in Table~\ref{tab: dataset}.
 
\begin{table*}[t]
\centering
\begin{tabular}{llllll}
\toprule
                 \textbf{Datasets}         &   \textbf{Model}    & \textbf{(Train Test)}                & \textbf{Response Number} & \textbf{ACS} & \textbf{ARC} \\
\midrule
\multirow{2}{*}{LongFact} & LLaMA-3 & \multirow{2}{*}{(2097,233)} & \multirow{2}{*}{30}      & 290                               & 23                                 \\
                          & Phi-3.5-mini   &                             &                          & 245                               & 14  \\ 
\bottomrule
\end{tabular}
\caption{Summary of the training dataset statistics. Response Number denotes the number of initial responses generated per question. ACS (Average Cluster Size) represents the average number of clusters formed per question based on atomic fact clustering. ARC (Average Response Coverage) indicates the average number of clusters that each response contributes to\label{tab: dataset}} 
\end{table*}

\subsection{Data Generation, Training, Evaluation Prompts}

The system prompt we use for the initial response generation is modified from FactTune~\citep{tian2023fine}, 
The User Prompt (questions) is kept the same as FactAlign~\citep{huang2024factalign}.
\begin{tcolorbox}[title=Initial Response Generation (Training Data Creation), colback=gray!15, colframe=black!70, boxrule=0.5pt]
\textbf{System Prompt:} \\
\textit{"You are an intelligent assistant who answers questions accurately."}
\vspace{1em}\\
\textbf{User Prompt:} \\
\textit{"What is the geographical importance of the Strait of Gibraltar? Provide as many specific details and examples as possible (such as names of people, numbers, events, locations, dates, times, etc)."}
\end{tcolorbox}

The training prompt for DPO alignment is kept as the default, which is the same question prompt as the generation part.

For the test data generation, 
The actual question exactly follows FactScore \citep{min2023factscore} official repository:
\begin{tcolorbox}[title=Test Response Generation, colback=gray!15, colframe=black!70, boxrule=0.5pt]
\textbf{System Prompt:} \\
\textit{"You are an intelligent assistant who answers questions accurately."}

\vspace{1em}
\textbf{User Prompt:} \\
\textit{"Answer this question. Question: Tell me a bio of Kourosh Zolani."}
\end{tcolorbox}

\end{document}